\documentclass[runningheads]{llncs}
\usepackage{graphicx}

\usepackage{tikz}
\usepackage{comment}
\usepackage{amsmath,amssymb} 
\usepackage{color}

\usepackage{bbm}
\usepackage{multirow}
\usepackage{arydshln}
\usepackage[export]{adjustbox}
\usepackage{soul}
\usepackage{booktabs}
\usepackage{pifont}
\usepackage[linesnumbered,ruled,vlined]{algorithm2e}
\usepackage{array}
\usepackage{enumitem}
\usepackage{caption}
\usepackage{subcaption}
\usepackage{wrapfig}
\usepackage[calcwidth = .9\linewidth, font = small, labelfont = bf]{caption}

\DeclareMathOperator*{\argmax}{arg\,max}

\newcommand{\regnas}{\textsc{REG-NAS}\xspace}
\newcommand{\eg}{\emph{e.g.}\xspace}
\newcommand{\ie}{\emph{i.e.}\xspace}
\newcommand{\vs}{\emph{v.s.}\xspace}
\newcommand{\etal}{\emph{et.al.}\xspace}

\usepackage[capitalize]{cleveref}
\crefname{section}{Sec.}{Secs.}
\Crefname{section}{Section}{Sections}
\Crefname{table}{Table}{Tables}
\crefname{table}{Tab.}{Tabs.}

\begin{document}
\pagestyle{headings}
\mainmatter
\def\ECCVSubNumber{4124}  

\title{Towards Regression-Free Neural Networks for Diverse Compute Platforms} 

\author{
Rahul Duggal\thanks{Currently at Georgia Tech. Work conducted during an internship with AWS AI.} \quad \ 
Hao Zhou  \quad \ 
Shuo Yang  \quad \\
Jun Fang  \quad \
Yuanjun Xiong  \quad \
Wei Xia 
}
\authorrunning{Duggal et al.}
\institute{AWS/Amazon AI \\
{\tt\small  rduggal7@gatech.edu} \quad
{\tt\small \{zhouho, shuoy, junfa, yuanjx, wxia\}@amazon.com} }
 \maketitle

With the shift towards on-device deep learning, ensuring a consistent behavior of an AI service across diverse compute platforms becomes tremendously important.
Our work tackles the emergent problem of reducing predictive inconsistencies arising as negative flips: test samples that are correctly predicted by a less accurate model, but incorrectly by a more accurate one.
We introduce \textbf{REG}ression constrained \textbf{N}eural \textbf{A}rchitecture \textbf{S}earch (\regnas) to design a family of highly accurate models that engender fewer negative flips.
\regnas consists of two components: (1) A novel architecture constraint that enables a larger model to contain all the weights of the smaller one thus maximizing weight sharing.
This idea stems from our observation that larger weight sharing among networks leads to similar sample-wise predictions and results in fewer negative flips;
(2) A novel search reward that incorporates both Top-1 accuracy and negative flips in the architecture search metric.
We demonstrate that \regnas can successfully find desirable architectures with few negative flips in three popular architecture search spaces.
Compared to the existing state-of-the-art approach \cite{Yan_2021_CVPR}, \regnas enables $33-48\%$ relative reduction of negative flips.
%

\section{Introduction}
\label{sec:intro}
\interfootnotelinepenalty=10000

\begin{figure}[h!]
    \centering
    \includegraphics[width=\linewidth]{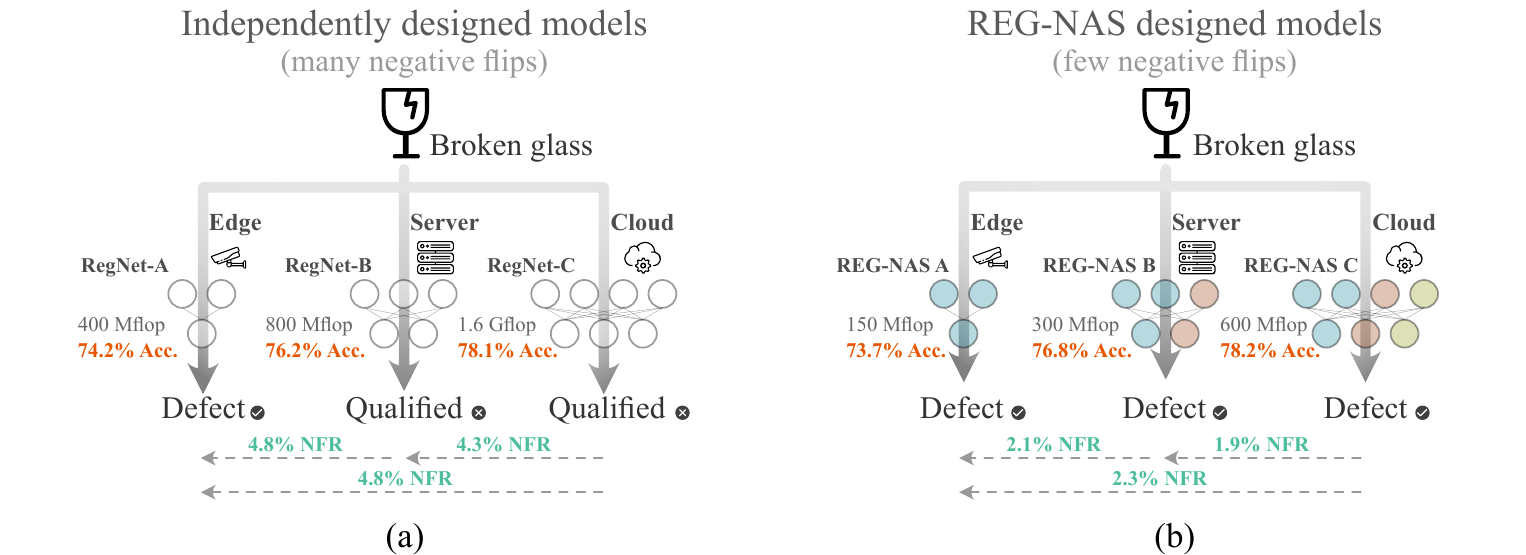}
    \vspace{-0.3cm}
    \caption{Regression arises when samples that are correctly predicted by a less accurate model are \textit{negatively flipped} or incorrectly predicted by a more accurate one. (a) Three models from the popular RegNet~\cite{RegNet} family trained on ImageNet suffer from $4.3-4.8\%$ pairwise negative flip rate (NFR); (b) Models designed via \regnas achieve $1.9-2.3\%$ NFR with similar Top-1 accuracy.}
    \vspace{-0.5cm}
    \label{fig:intro_crown_jewel}
\end{figure}

Consider a manufacturer that uses deep learning to detect defective products on their assembly line.
To improve throughput, they use a small, low-latency, on-device model to quickly detect potentially defective products which later undergo a secondary examination by a large, and more accurate model deployed on the cloud.
If there are many cases where the on-device model correctly identifies a defective product which, on second inspection the on-cloud model incorrectly believes to be qualified, the defective product may end up entering the market and damaging the reputation of the manufacturer (See Fig.~\ref{fig:intro_crown_jewel}a).

Such samples, which one model predicts correctly while the other predicts incorrectly, are called negative flips.
The fraction of negative flips over the dataset size is called the negative flip rate (NFR) \cite{Yan_2021_CVPR} and is used to measure regression between two models\footnote{Please note this definition of regression is different from the traditional one which pertains to predicting the outcome of continuous variables.}.
In this work, we aim to design a family of models for diverse compute budgets (\eg deployed on edge, server, and cloud) that maximize Top-1 accuracy and minimizes the pairwise negative flip rate (See Fig.~\ref{fig:intro_crown_jewel}b).

Positive congruent training \cite{Yan_2021_CVPR} is the first work to tackle negative flips.
They propose the Focal Distillation (FD) loss, which enforces congruence with the reference model by giving more weights to samples that were correctly classified by the reference model.
While \cite{Yan_2021_CVPR} aims at reducing flips along the \textit{temporal} axis, \ie between different model versions trained on growing datasets, we tackle regression along the \textit{spatial} axis, \ie between models deployed on diverse compute platforms.
The key difference being that the temporal setting constrains the model design to necessarily be \textit{sequential} since the past model is fixed while the current model is designed and the future model is unknown.
In contrast, the spatial setting enjoys larger flexibility wherein all models can be \textit{jointly} adapted towards minimizing negative flips.

\begin{figure}[t!]
    \centering
    \includegraphics[width=0.95\linewidth]{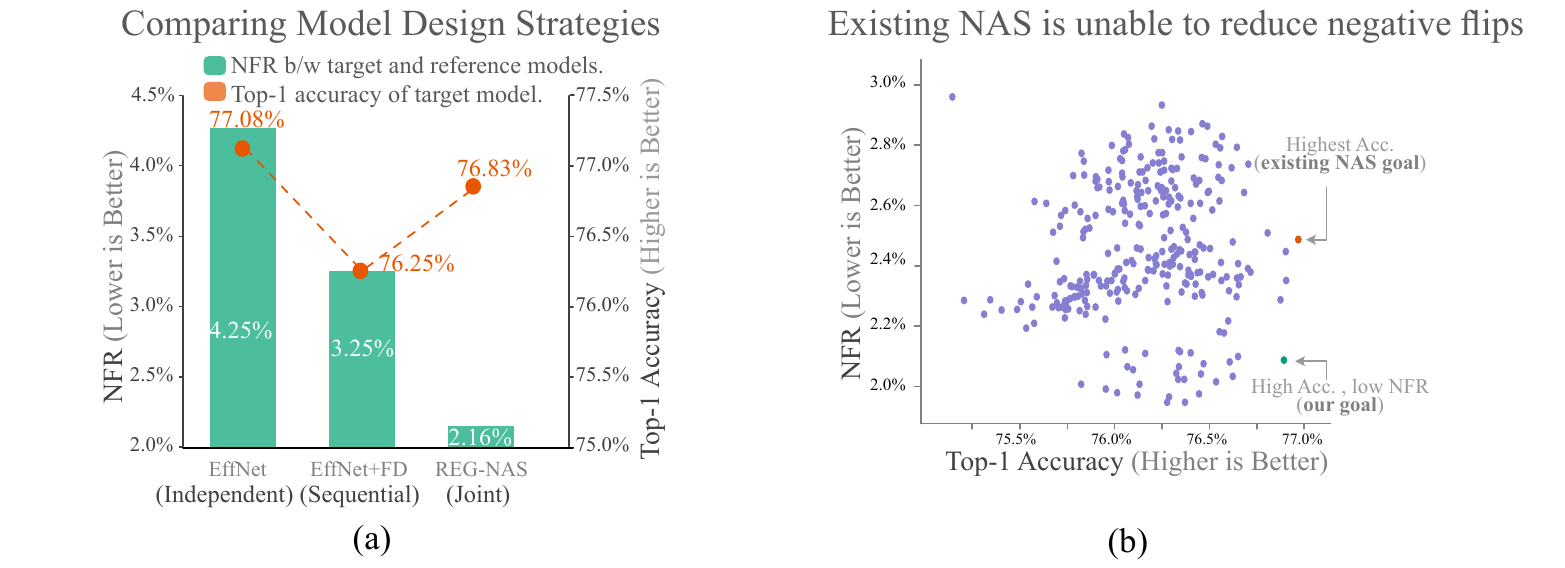}
    \caption{(a) Independently training a MobileNet-V3 and EfficientNet-B1 as reference and target models leads to 73.6\%, 77.08\% Top-1 accuracies with 4.25\% negative flip rate (NFR). Sequentially training the target against the reference model with the Focal Distillation loss~\cite{Yan_2021_CVPR} reduces NFR to the detriment of target model's accuracy. Jointly designing the reference and the target models via \regnas leads to lowest NFR with negligible drop in accuracy. (b) Plotting the Top-1 accuracy and NFR (measured against a common reference model) for 300 randomly sampled architectures. Observe that the NFR and Top-1 accuracy are uncorrelated which means existing NAS that maximizes Top-1 accuracy may fail to find models with the lowest negative flip rate.}
    \vspace{-0.5cm}
    \label{fig:architecture_impact_on_nfr}
\end{figure}

To quantify the difference between sequential and joint design strategies, Fig.~\ref{fig:architecture_impact_on_nfr}a presents the case of designing a small (reference) and a large (target) model.
Independently training a MobileNet-v3 reference and an EfficientNet-B1 target model using cross entropy loss achieves 73.6\% and 77.08\% Top-1 accuracies with more than 4.25\% negative flip rate.
Training the target model \textit{sequentially} against the fixed reference model via focal distillation loss~\cite{Yan_2021_CVPR} reduces the NFR to $3.25\%$ with a $0.8\%$ drop in Top-1 accuracy. 
However, \textit{jointly} designing the reference and target models using \regnas significantly reduces the NFR to $2.16\%$ with a marginal $0.25\%$ drop in the Top-1 accuracy.

\regnas can jointly design models by sampling sub-networks from a common super-network.
We hypothesize this implicitly reduces negative flips due to large weight sharing among the sampled sub-networks (see appendix A for a detailed study).
This naturally motivates the question: Could we further reduce negative flipping by maximizing weight sharing?
To this end, we propose the first component of \regnas---a novel \textit{architecture constraint} that maximizes weight-sharing by enabling the larger sub-networks to contain all the weights of smaller ones.
Empirically, in Sec.~\ref{subsec:architecture_constraint} we show that architectures satisfying this constraint exhibit much lower negative flips while still achieving high accuracy.

While weight sharing implicitly helps reduce negative flips, it is not sufficient.
Fig.~\ref{fig:architecture_impact_on_nfr}b shows that among architectures that maximally share weights by satisfying the architecture constraint, there is a significant variance in NFR and Top-1 accuracy. 
To search for highly accurate architectures that minimize regression, in Sec.~\ref{subsec:search_reward} we propose the second component of \regnas---a novel \textit{search reward} that includes both Top-1 accuracy and NFR as the architecture search metric.

Together, our proposed architecture constraint and novel search reward enable \regnas to reduce negative flips by $33-48\%$ over the existing state-of-the-art FD loss~\cite{Yan_2021_CVPR}. 
We also show that \regnas can find low NFR sub-networks from popular architecture search spaces, demonstrating its  strong generalization ability.
Additionally, the models searched via \regnas are transitive under the NFR metric, enabling the search for N ($\geq2$) models across diverse compute budgets.
To summarize, our contributions are: 
\begin{enumerate}
    \item We extend the search for regression-free models from the temporal to the spatial setting. This support scenarios that simultaneously deploy a family of models across diverse compute budgets.
    \item  We propose two novel ideas to minimize negative flips while achieving high Top-1 accuracy: an architecture constraint that maximizes weight sharing thereby inducing similar sample-wise behavior; and a novel architecture search reward to explicitly include NFR in the architecture search metric. 
    \item We comprehensively evaluate our search method on ImageNet and demonstrate its generalization on three popular architecture search spaces.
\end{enumerate}
\section{Related Work and Background}
\noindent \textbf{Minimizing regression} between different models has previously been explored in the temporal setting where different model versions arise during model updates~\cite{BCT,srivastava2020empirical}. 
Generally, this task closely relates to research in continual learning~\cite{kirkpatrick2017overcoming,chen2018lifelong,parisi2019continual}, incremental learning~\cite{castro2018end,wu2019large,belouadah2019il2m} and model compatibility~\cite{BCT,Yan_2021_CVPR,CMPNAS}. 
However, the main difference lies in how regression is measured. 
Traditionally, regression or forgetting was measured via the error rate~\cite{lopez2017gradient,li2017learning,chaudhry2018riemannian}, however, Yan \etal \cite{Yan_2021_CVPR} observe that two models having the similar error rates may still suffer from a high negative flip rate.
Moreover, the methods that proposed to alleviate catastrophic forgetting are insufficient at reducing NFR.
Different from these works, our work tackles NFR in the spatial setting wherein different models arise due to a  diversity in deployment targets (\eg edge, server, cloud).
The spatial setting is fundamentally different from the temporal one since it enjoys the additional flexibility of jointly designing all architectures that can benefit from weight sharing that ultimately leads to similar sample-wise behavior and lower NFR.
This key distinction calls for new methods that can leverage the flexibility of joint architecture design.

\medskip
\noindent \textbf{Neural architecture search} (NAS) is a popular tool for automated design of neural architectures. 
Earlier NAS methods based on reinforcement learning~\cite{zoph2016neural,liu2017hierarchical,tan2019mnasnet}, evolutionary algorithms~\cite{xie2017genetic,real2017large,suganuma2018exploiting,real2019regularized}, and Bayesian optimization~\cite{kandasamy2018neural} find highly accurate networks but are notoriously computational heavy due to the independent evaluation of candidate networks. 
Recently One-Shot NAS approaches~\cite{pham2018efficient,DARTS,yu2019autoslim} successfully adopt weight sharing to alleviate the compute burden. 
The main innovation in this area is a super-network~\cite{guo2020single,OFA,BIGNAS,AttentiveNAS,alphanet,dai2021fbnetv3,wu2021fbnetv5} that can encode and jointly
 train all candidate sub-networks in an architecture space. 
Typically, One-Shot NAS algorithms have the following two phases:

 \textbf{[P1] Super-network pre-training:} The goal of which is to jointly optimize all sub-networks through a weight sharing super-network.
The optimization process typically samples and minimizes the loss on many individual sub-networks during training.
Recent methods innovate on the sampling strategy~\cite{OFA,BIGNAS,AttentiveNAS,dai2021fbnetv3,wu2021fbnetv5} or the loss formulation~\cite{alphanet}.

\textbf{[P2] Sub-network searching:} The goal of this phase is to find the best sub-network from the well-trained super-network under some compute measure such as flops or latency.
This search is typically implemented via an evolutionary algorithm~\cite{OFA,AttentiveNAS,alphanet} with Top-1 accuracy as the reward function.

Our work leverages a super-network for an entirely new benefit-- for reducing negative flips between networks optimized for diverse compute platforms. 
This is yet an unexplored problem that we believe will become increasingly important in the era of on-device AI powered by hardware optimized neural networks.

\section{Regression Constrained Neural Architecture Search}
Our \regnas algorithm relates to the super-network searching phase (\textbf{P2}) and generally works with any super-network obtained via the recent pre-training strategies (\textbf{P1})~\cite{OFA,AttentiveNAS,alphanet}.
We modify the search optimization of \textbf{P2} along two directions:
We propose a novel architecture constraint that maximizes weight sharing among sub-networks which is discussed in Sec.~\ref{subsec:architecture_constraint}.
We propose a novel search reward that incorporates both Top-1 accuracy and the NFR metric as described in Sec.~\ref{subsec:search_reward}.
Finally, we integrate these two ideas into our \regnas algorithm as discussed in Sec.~\ref{subsec:regnas}.

\subsection{Measuring regression}
\label{subsec:nfr}
We use the negative flip rate (NFR) \cite{Yan_2021_CVPR} as the metric to measure the regression between two models.
Given a dataset $\mathcal{D}$ that contains image and label pairs $(x_i, y_i)$, $i\in \{1, ..., N\}$, the output of two deep CNNs $\phi_1, \phi_2$ on an input image $x_i$ are denoted by $y_i^1$ and $y_i^2$.
The NFR between $\phi_1, \phi_2$ is defined as:
\begin{equation}
\label{eq:nfr_def}
\text{NFR}(\phi_1, \phi_2; \mathcal{D}) = \frac{1}{N}\sum_{i=1}^{N}\mathbbm{1}(y_i^1 = y_i, y_i^2 \neq y_i),
\end{equation}
where $\mathbbm{1}$ is an indicator function.
The NFR  measures the fraction of samples that are predicted correctly by $\phi_1$ and incorrectly by $\phi_2$. Note that NFR is asymmetric, so $\text{NFR}(\phi_1, \phi_2) \neq \text{NFR}(\phi_2, \phi_1)$.
We select $\phi_1$ as the model with a lower Top-1 accuracy and $\phi_2$ as the one with the higher Top-1 accuracy.
As a result, $\text{NFR}(\phi_1, \phi_2) \in [0,1]$.

\subsection{Architecture constraint to reduce negative flips}
\label{subsec:architecture_constraint}

We hypothesize that sub-networks sampled from a super-network share a lot of weights, which induces similar sample-wise behavior and results in less negative flips.
A natural extension of this hypothesis prompts the question: can we further reduce negative flips by maximizing weight-sharing between sub-networks?
In what follows, we describe the architecture constraint that answers this question.

Consider a reference model with architecture $a_r$ that inherits weights $W^*_{a_r}$ from a super-network.
We aim to search for a target architecture $a_t$ with weights $W^*_{a_t}$ such that the weight sharing between $a_r$ and $a_t$ is maximized, \ie:
\begin{align}
&a_t\ = \argmax_{a \in \mathcal{A}} \mathcal{N}(W^*_{a} \cap W^*_{a_r}),  \label{eq:constrained_search}\\
& \text{s.t.}~~C(a) < \tau, \nonumber
\end{align}
where $\mathcal{N}$ counts the number of weights, $\cap$ represents the intersection of weights between two architectures, $C$ represents a performance constraint such as flops or latency and $\mathcal{A}$ represents the search space defined by the super-network.
Eq.~\ref{eq:constrained_search} specifies the architecture constraint and we refer to the process of solving it as \textbf{constrained architecture search (CAS)}.

Observe that $N(W^*_{a} \cap W^*_{a_r})$ is upper bounded by $N(W^*_{a_r})$, with the supremum occurring when $a$ contains all the weights of $a_r$.
This means Eq.~\ref{eq:constrained_search} partitions the architecture space $\mathcal{A}$ to a space $\mathcal{A}_t$ of constrained architectures:
\begin{align}
\mathcal{A}_t = \{a: W^*_{a_r} \subseteq W^*_{a}, \forall a \in \mathcal{A}\},
\label{eq:constraiend_architecture_space}
\end{align}
and its complementary space $\mathcal{A}_c = \mathcal{A} - \mathcal{A}_t$ that includes target architectures that do not contain \textit{all} the  weights of the reference model. 
%
 To visualize the two architecture spaces,  we plot NFR \vs Top-1 accuracy for 300 randomly sampled architectures from $\mathcal{A}_t$ and $\mathcal{A}_c$ in Fig.~\ref{subfig:architecture_space}.
Observe that the architectures in space $\mathcal{A}_t$ generally have lower NFR and higher Top-1 accuracy compared to those in $\mathcal{A}_c$.
As a result, it is easier to find architectures with lower NFR and higher Top-1 accuracy in space $\mathcal{A}_t$ than in the complete space $\mathcal{A}$ as demonstrated empirically in Sec~\ref{subsec:ablation}.

\begin{figure}[!h]
    \centering
    \vspace{-0.5cm}
    \includegraphics[width=\textwidth]{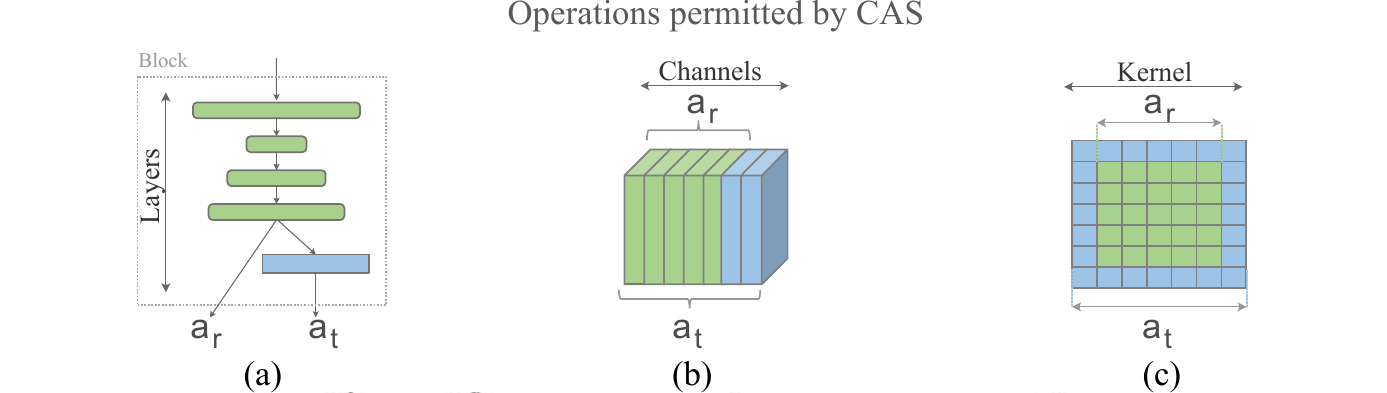}
    \caption{
      Constrained Architecture Search (CAS) recipe specifies three operations for searching a target model $a_t$ given a reference model $a_r$: (a) Adding new layers within an existing block of $a_r$; (b) Adding new channels to existing filters in $a_r$; (c) Extending the kernel size beyond the existing one in $a_r$. 
    }
    \vspace{-0.5cm}
    \label{fig:CSR}
\end{figure}

Note that, not all super-networks allow for partitioning the search space according to Eq.~\ref{eq:constraiend_architecture_space}.
For example DARTS~\cite{DARTS,FAIR_DARTS} implements different operators (\eg $3\times3$ and $5\times5$ conv) using a completely non-overlapping set of weights.
In such search spaces, enforcing the architecture constraint trivially leads to the same architecture for the reference and target model.  
On the other hand, recent works~\cite{OFA,AttentiveNAS,alphanet} implement different operators using an overlapping set of weights \eg the innermost $3\times3$ weights of the $5\times5$ kernel implements a $3\times3$ kernel.
To enable the architecture constraint for such works, we propose the CAS recipe which specifies three rules for searching a candidate architecture $a_t$ against a reference model $a_r$:
\begin{enumerate}[itemsep=0em]
    \item Add new layers within blocks of $a_r$ (Fig.~\ref{fig:CSR}a).
    \item Add additional channels to filters in $a_r$ (Fig.~\ref{fig:CSR}b).
    \item Extend kernel size for filters in $a_r$ (Fig.~\ref{fig:CSR}c).
\end{enumerate}

\subsection{Search reward to reduce NFR}
\label{subsec:search_reward}
\begin{figure*}
    \centering
    \begin{subfigure}{0.46\linewidth}
    \centering
    \includegraphics[width=\linewidth]{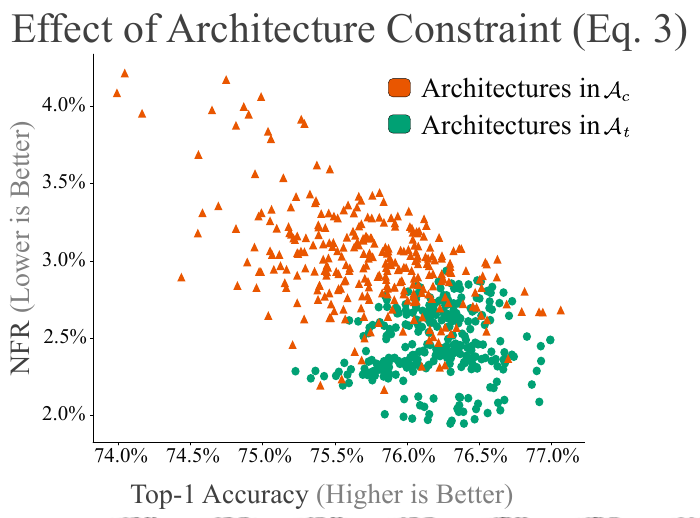}
    \caption{}
    \label{subfig:architecture_space}
    \end{subfigure}
     \begin{subfigure}{0.46\linewidth}
    \centering
    \includegraphics[width=\linewidth]{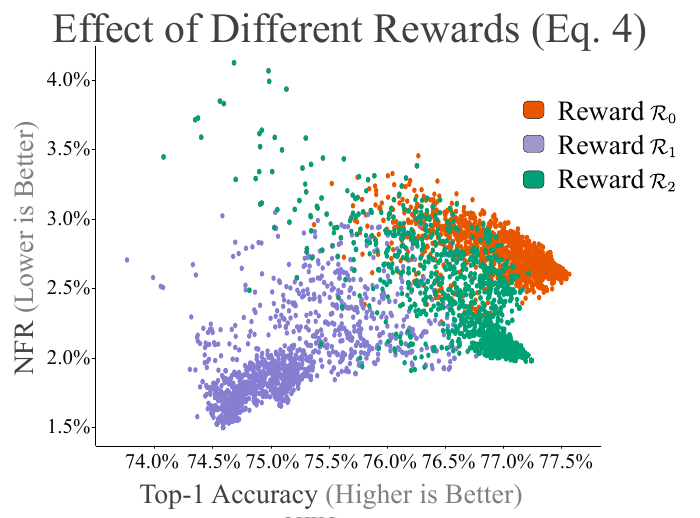}
    \caption{}
    \label{subfig:search_reward_impact}
    \end{subfigure}
    \caption{Through random sampling architectures from the MobileNet-V3 search space $\mathcal{A}$ of OFA~\cite{OFA}, we show: (a) The architecture constraint partitions $\mathcal{A}$ into $A_t$ and $A_c$ wherein the architectures in $A_t$ tend to have lower NFR; (b) An evolutionary search fitted with  rewards ($\mathcal{R}_0, \mathcal{R}_1, \mathcal{R}_2$) explores different regions of the architecture space. Specifically, $\mathcal{R}_0$ leads to architectures with high Top-1 and NFR, $\mathcal{R}_1$ leads to architectures with low Top-1 and NFR while $\mathcal{R}_2$ achieves the best of both--high Top-1 accuracy and low NFR.}
    \label{fig:nas_motivation}
\end{figure*}

While the architecture constraint implicitly helps reduce NFR through weight sharing, it is not sufficient.
To illustrate this, we plot the Top-1 accuracy and the NFR of 300 randomly sampled sub-networks from $\mathcal{A}_t$ against a fixed reference model in Fig.~\ref{fig:architecture_impact_on_nfr}b. 
Observe that the model with highest Top-1 accuracy has a NFR around $2.5\%$ which is higher than many of the sampled architectures.
This means that optimizing solely for the Top-1 accuracy as in existing NAS methods will not lead to architectures with small NFR.
Consequently, to optimize for architectures with low NFR it is essential to include NFR in the search reward.

The observations of Fig.~\ref{fig:architecture_impact_on_nfr}b naturally motivate the inclusion of NFR in the search reward $\mathcal{R}$ as below: 
\begin{eqnarray}
    \mathcal{R}\left(W^*_{a_t}; W^*_{a_r},~\mathcal{D}^{val}\right) &=& \lambda_1 * \text{Top-1}(W^*_{a_t}; ~\mathcal{D}^{val}) \\
    &-& \lambda_2 * \text{NFR}(W^*_{a_t}, W^*_{a_r}; \mathcal{D}^{val}), \nonumber
    \label{eq:our_reward}
\end{eqnarray}
where $W^*_{a_r}, W^*_{a_t}$ represents the weights of the reference and target models while $\lambda_1, \lambda_2$ balance the trade off between Top-1 accuracy and NFR respectively.
To illustrate the extreme effect of the two terms, we consider three different rewards obtained by setting the following values of $\lambda_1$ and  $\lambda_2$.
\begin{enumerate}[noitemsep]
    \item[$\mathcal{R}_0$] ($\lambda_1=1, \lambda_2=0$): reward used by existing NAS methods that solely optimizes for high Top-1 accuracy.
    \item[$\mathcal{R}_1$] ($\lambda_1=0, \lambda_2=1$): reward that solely optimizes for low NFR.
     \item[$\mathcal{R}_2$] ($\lambda_1=1, \lambda_2=1$): reward that optimizes for high accuracy and low NFR.
\end{enumerate}

We find that the rewards $\mathcal{R}_0$, $\mathcal{R}_1$ and $\mathcal{R}_2$ can guide evolutionary search to different regions of the architecture search space.
Fig.~\ref{subfig:search_reward_impact} illustrates this by plotting the Top-1 accuracy and NFR of all architectures sampled during an evolutionary search fitted with the three rewards.
The plot clearly shows that $\mathcal{R}_2$ guides the search to architectures with high Top-1 and low NFR.
We consider $\mathcal{R}_2$ as the default choice for \regnas since it equally balances the Top-1 accuracy and NFR. 
Please refer to appendix B for a complete ablation on $\lambda_1, \lambda_2$.

\subsection{\regnas: Regression constrained neural architecture search}
\label{subsec:regnas}
To search for sub-networks that achieve the highest Top-1 accuracy with the minimum negative flips, we integrate the novel architecture constraint of Eq.~\ref{eq:constraiend_architecture_space} with the novel reward of Eq.~\ref{eq:our_reward} to propose \regnas, which solves the following optimization.
\begin{align}
&a_t = \argmax_{a \in \mathcal{A}}~~  \mathcal{R}\left(W^*_{a}; W^*_{a_r},~\mathcal{D}^{val}\right),  \label{eq:supernet_search}\\
& \text{s.t.}~~W^*_{a_r} \subseteq W^*_{a},  \text{and}~~C(a) < \tau \nonumber
\end{align}

We solve the above optimization using a standard evolutionary based NAS algorithm with two modifications: the random sampling and mutate operations can only perform the operations specified by the CAS in Sec.~\ref{subsec:architecture_constraint}; and the search reward uses the reward formulation of Sec.~\ref{subsec:search_reward} with $\lambda_1=1, \lambda_2=1$.

\newcommand\mycommfont[1]{\textbf{\footnotesize\ttfamily\textcolor{blue}{#1}}}
\SetCommentSty{mycommfont}

\section{Experiments}
We begin by comparing \regnas against the state-of-the-art method \cite{Yan_2021_CVPR} designed to reduce NFR.
To understand the effectiveness of \regnas, we dissect the performance gains due to the novel architecture constraint and search reward.
We then extend the proposed method to search for an entire family of architectures.
Finally, we test the generalization of \regnas across: different flop budgets, super-networks, and latency constraints on diverse platforms.

\subsection{Dataset, metrics and implementation details}
We implement our methods using Pytorch on a system containing 8 V100 GPUs. Other details are as follows:

\smallskip \noindent \textbf{Dataset.} We present all results on ImageNet \cite{imagenet}.
We carve out a sub-training set of 20,000 images from the original training set to evaluate the rewards and update batchnorm statistics for each sub-network~\cite{OFA}.
The final results are presented on the original ImageNet validation set.

\smallskip \noindent \textbf{Evolutionary Search.} To search for a sub-network, we run the evolutionary search with the following hyper-parameters: 20 generations,  100 population size, 0.1 mutate probability, 0.5 mutation ratio and 25\% Top-K architectures moving into the next generation.
Overall, our search evaluates 1,525 architectures in each super-network.
Following the setting of the original work, for MobileNet-V3 and ResNet-50 search spaces of OFA~\cite{OFA}, the rewards are computed on the sampled sub-training set, while for FBNet-V3 super-network of AttentiveNAS~\cite{AttentiveNAS} and AlphaNet~\cite{alphanet}, the search is performed on the ImageNet validation set as suggested by \cite{AttentiveNAS,alphanet}.

\smallskip \noindent \textbf{Notation.} We refer to different models following the consistent notation AA-BB-CC. The first initial ``AA'' refer to the super-network, the middle initial ``BB'' refers to the search reward and the final initial ``CC'' refers to the performance constraint such as flops or latency. 
We use CAS to represent constrained architecture search discussed in Sec.~\ref{subsec:architecture_constraint}.

\subsection{Regression Constrained Search}
\label{subsec:regression_constrained_search}

Given a small reference model, we aim to find a larger target model that, under a fixed compute budget, maximizes the Top-1 accuracy while minimizing NFR.
Table~\ref{tab:main_results} illustrates this task with comparative results from the MobileNet-V3 and ResNet-50 search spaces of OFA~\cite{OFA}.

\begin{table}[h!]
    \centering
    \begin{subtable}{0.49\textwidth}
        \begin{adjustbox}{width=0.95\columnwidth}
        \large
        \begin{tabular}{llccc}
            \toprule
            & Model & Flops  & Top-1  & NFR \\
            &  &   & $\uparrow$(\%) & $\downarrow$ (\%) \\
            \midrule
            ref & MB-$\mathcal{R}_0$-150~\cite{OFA}  & 149 & 73.70 & - \\ \cdashline{1-5}[.7pt/1.5pt]\noalign{\vskip 0.15em}
            baseline  & EffNet     & 385 & 77.08  & 4.25\\
            PCT  & EffNet+FD~\cite{Yan_2021_CVPR}  & 385 & 76.25  & 3.25\\
            wt. share & MB-$\mathcal{R}_0$-300~\cite{OFA}  & 298 & 77.11  & 2.51\\
            \regnas &  MB-($\mathcal{R}_2$+CAS)-300 & 294 & 76.83 & 2.16 \\
            \bottomrule
        \end{tabular}
      \end{adjustbox}
      \caption{MobileNet-V3 search space of OFA}
      \label{subtab:main_results_mbv3}
    \end{subtable}
    \begin{subtable}{0.49\textwidth}
    \begin{adjustbox}{width=0.95\columnwidth}
        \large
        \begin{tabular}{llcccc}
            \toprule
          & Model & Flops  & Top-1  & NFR \\
            &  &   & $\uparrow$(\%) & $\downarrow$ (\%) \\
            \midrule
            ref & RN-$\mathcal{R}_0$-2000~\cite{OFA}  & 1973 & 78.25 & - \\ \cdashline{1-5}[.7pt/1.5pt]\noalign{\vskip 0.15em}
            baseline & RN101 & 7598 & 79.21 & 4.83 \\
            PCT     & RN101+FD~\cite{Yan_2021_CVPR}  & 7598 & 79.90 & 3.06 \\
            wt. share & RN-$\mathcal{R}_0$-3000~\cite{OFA}  & 2941  & 78.78 & 2.04 \\
            \regnas &  RN-($\mathcal{R}_2$+CAS)-3000 & 2915 & 78.80 & 1.57 \\
            \bottomrule
        \end{tabular}
      \end{adjustbox}
      \caption{ResNet search space of OFA}
      \label{subtab:main_results_rn50}
    \end{subtable}
    \caption{
    Comparing different model design strategies. We present results for MobileNet-V3 (MB) and ResNet-50 (RN) search spaces of OFA~\cite{OFA}. EffNet and RN101 represents EfficientNet-B0 and ResNet-101 trained with cross entropy loss, while
     EffNet+FD and RN101+FD are trained with state-of-the-art focal distillation loss \cite{Yan_2021_CVPR}.
     Results of weight sharing and our proposed method are averaged from three runs with different random seeds.}
     \vspace{-0.5cm}
    \label{tab:main_results}
\end{table}

We first search for the best reference model under $150$ mega flops constraint (MB-$\mathcal{R}_0$-150) using the $\mathcal{R}_0$ reward which maximizes Top-1 accuracy.
Then, we establish a baseline by training an off-the-shelf EfficientNet-B0 model\footnote{We choose EfficientNet-B0 since it has similar Top-1 accuracy with sub-networks searched from MobileNet-V3 in OFA \cite{OFA}} using vanilla cross entropy. 
This model improves Top-1 accuracy to $77\%$, but suffers from 4.25\% negative flips.
A stronger baseline can be achieved by training the EfficientNet using Focal Distillation~\cite{Yan_2021_CVPR} which is the current state-of-the-art method to reduce negative flips. 
It reduces the NFR to $3.25\%$ with a $0.8\%$ drop in Top-1 accuracy.
Another strong weight sharing baseline can be obtained via searching a sub-network from the same super-network as the reference model using $\mathcal{R}_0$ reward. This leads to an NFR of $2.51\%$ with $77.11\%$ Top-1 accuracy. 
Compared to all of these, the networks obtained via \regnas \ie (MB-($\mathcal{R}_2$+CAS)-300) achieves the lowest NFR of $2.16\%$ with less than $0.3\%$ drop in Top-1 accuracy compared with MB-$\mathcal{R}_0$-300. 
This constitutes a $50\%$ and $33\%$ relative reduction of NFR with respect to the baseline and existing state-of-the-art method with less than $1\%$ change in accuracy (see appendix C).
The strong performance of \regnas  clearly demonstrates the benefits of the search reward and architecture constraint.
Similar results can be observed for the ResNet-50 search space of OFA~\cite{OFA} in Table~\ref{subtab:main_results_rn50}.

\begin{table}[t]
    \centering
    \begin{subtable}[t]{0.49\textwidth}
        \begin{adjustbox}{width=0.95\columnwidth}
        \huge
        \begin{tabular}{llccc}
            \toprule
             & Model & Flops  & Top-1  & NFR  \\
              &   &  & $\uparrow$ (\%) & $\downarrow$ (\%)  \\
            \midrule
             ref & MB-$\mathcal{R}_0$-150 &  149 & 73.70 & -  \\ \cdashline{1-5}[.7pt/1.5pt]\noalign{\vskip 0.15em}
             paragon Top-1 & MB-$\mathcal{R}_0$-300  & 298 & 77.11 & 2.51\\
             paragon NFR & MB-$\mathcal{R}_1$-300  & 279 & 76.20 & 2.07  \\
             & MB-$\mathcal{R}_2$-300  & 292 & 76.90 & 2.25  \\
            \regnas & MB-($\mathcal{R}_2$+CAS)-300 & 294 & 76.83 & 2.16  \\
            \bottomrule
        \end{tabular}
       \end{adjustbox}
       \hspace{1em}
       \caption{MobileNet-V3 search space of OFA}
       \hspace{2em}
       \label{subtab:rewards_ablation_mbv3}
    \end{subtable}
    \begin{subtable}[t]{0.49\textwidth}
    \begin{adjustbox}{width=0.95\columnwidth}
        \huge
        \begin{tabular}{llccc}
            \toprule
             & Model & Flops  & Top-1  & NFR \\
             &  &   & $\uparrow$(\%) & $\downarrow$ (\%) \\
            \midrule
            ref & RN-$\mathcal{R}_0$-2000 &   1973 & 78.25  & - \\ \cdashline{1-5}[.7pt/1.5pt]\noalign{\vskip 0.15em}
            paragon Top-1 & RN-$\mathcal{R}_0$-3000  & 2941 & 78.78 & 2.04\\
            paragon NFR & RN-$\mathcal{R}_1$-3000  & 2537 & 78.57 & 1.28 \\
             & RN-$\mathcal{R}_2$-3000  & 2972 & 78.87 & 1.76 \\
             \regnas & RN-($\mathcal{R}_2$+CAS)-3000  & 2915 & 78.80 & 1.57\\
            \bottomrule
        \end{tabular}
       \end{adjustbox}
       \caption{ResNet search space of OFA}
       \label{subtab:rewards_ablation_rn50}
    \end{subtable}
    \vspace{-0.5cm}
    \caption{
    Comparing different search rewards using the MobileNet-V3 (MB) and ResNet-50 (RN) search spaces of OFA~\cite{OFA}. Results (except ref) are averaged from three runs with different seeds.}
    \label{tab:rewards_ablation}
    \vspace{-1cm}
\end{table}

\subsection{Dissecting the performance of \regnas}
\label{subsec:ablation}
In this subsection, we study the contribution of different components of \regnas---the search reward and the constrained architecture search, in reducing NFR.
Table~\ref{tab:rewards_ablation} presents the performance of architectures searched via $\mathcal{R}_0$, $\mathcal{R}_1$, $\mathcal{R}_2$ and $\mathcal{R}_2$+CAS.
We observe that since $\mathcal{R}_0$ only includes Top-1 accuracy in its formulation, the optimal architecture searched via it achieves the best accuracy, which we treat as the paragon for Top-1 accuracy.
However, this same architecture also suffers from the highest NFR.
On the other hand, $\mathcal{R}_1$, which only includes the NFR in its formulation, leads to a model with the lowest NFR, and also lowest Top-1 accuracy. 
We treat this model as the paragon for NFR.

\noindent \textbf{Effect of reward $\mathcal{R}_2$.} By including both the Top-1 accuracy and NFR in its formulation, reward $\mathcal{R}_2$ leads to models that lie closer to the paragon of both metrics.
The best results, however, are achieved by including the architecture constraint which further brings the models closer to the respective paragons.
In the case of MobileNet-V3, architecture searched by $\mathcal{R}_2$+CAS is within $0.09\%$ of the paragon of NFR and within $0.28\%$ of the paragon of Top-1 accuracy, which demonstrates the effectiveness of the proposed method.

\begin{figure}[h!]
    \centering
    \includegraphics[width=\linewidth]{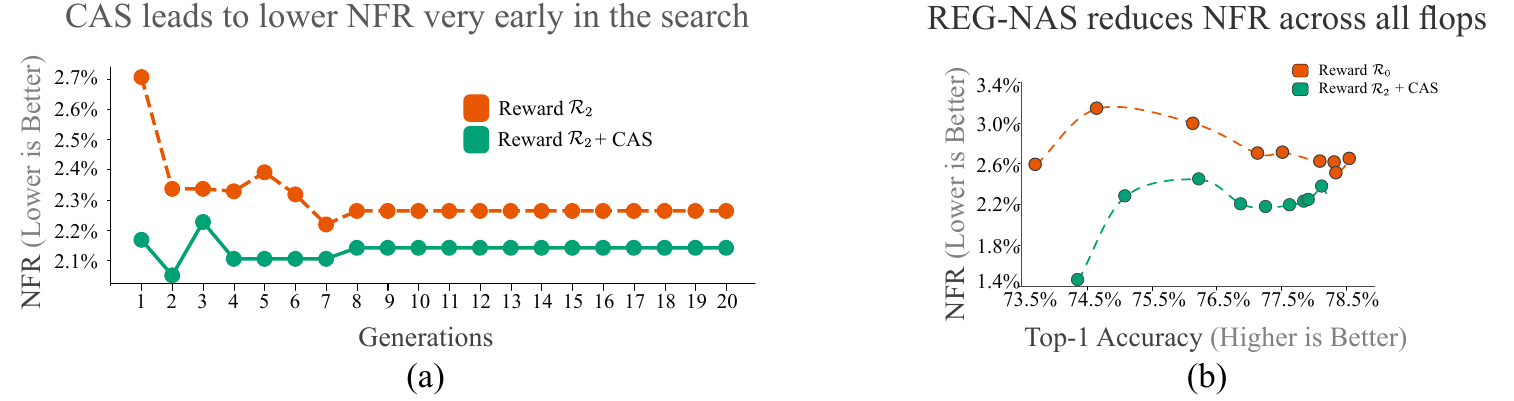}
    \vspace{-0.5cm}
    \caption{
    (a) NFR of the optimal architecture in each generation of the evolutionary search. Observe that CAS leads to lower NFR early and throughout the search. (b) Top-1 accuracy and NFR for 9 models (175-500 mflops) searched against a common reference (150 mflops). Observe that a smaller accuracy gap doesn't imply a smaller NFR. Also $\mathcal{R}_2+CAS$ outperforms $\mathcal{R}_0$ for all sizes.
    }
    \vspace{-0.5cm}
    \label{fig:nfr_convergence}
\end{figure}

\noindent \textbf{Effect of architecture constraint.} Besides reducing NFR, we find that CAS also help search for an architecture with low NFR much faster compared with $\mathcal{R}_2$ as shown in Fig.~\ref{fig:nfr_convergence}a.
This 
is because CAS partitions the overall architecture space into a subspace containing well performing architectures that achieve low NFR and high Top-1 accuracy as described in Sec.~\ref{subsec:architecture_constraint}.

\noindent \textbf{Effect of model size.} It is natural to ask if a smaller gap in Top-1 accuracy would imply a smaller NFR? The answer is provided in Fig.~\ref{fig:nfr_convergence}b which  plots the accuracy and NFR for 9 target models (from 175 to 600 Mflops) searched against a common 150 Mflop reference model with 73.7\% Top-1 accuracy.
Observe that with $\mathcal{R}_0$, even for the smallest model with similar accuracy as the reference, the NFR is consistently high ($\approx 2.5\%$).
This suggests that accuracy gap between reference and target has little impact on the NFR.
Additionally, we see that the search with $\mathcal{R}_2+CAS$ outperforms $\mathcal{R}_0$ across all model sizes.

\subsection{Supercharging \regnas with PCT}
We study if fine-tuning with FD loss proposed in \cite{Yan_2021_CVPR} can help further reduce the NFR for architectures searched via \regnas.
Table~\ref{tab:finetune} compares two loss functions: vanilla cross entropy (CE) and FD\footnote{We use FD-KD from \cite{Yan_2021_CVPR} since it outperforms FD-LM in our setting.} for fine-tuning the target sub-networks obtained via \regnas.

We observe that fine tuning with CE generally results in a higher NFR.
This is expected because CE optimizes the weights for higher Top-1 accuracy, which may push the weights of the target model away from those of the reference.
In contrast, FD loss promotes similar sample-wise behavior by penalizing negative flips and leads to a lower NFR.
This study demonstrates that FD loss can work in tandem with \regnas to further reduce NFR.
However, even without fine tuning, the architectures searched by \regnas already enjoy a very low NFR, even outperforming the MB-$\mathcal{R}_0$-300 architecture with fine tuning.
Considering the large time cost of fine tuning,
we avoid it in the following discussion.

\begin{table}[h!]
    \centering
        \begin{adjustbox}{width=0.6\columnwidth}
        \begin{tabular}{lllcc}
        \toprule
        Super-networks & Model      & Fine          & Top-1             & NFR\\
            &       & tune        & $\uparrow$(\%)    & $\downarrow$ (\%) \\
        \midrule
       \multirow{7}{*}{MobileNet-V3} & MB-$\mathcal{R}_0$-150 & - & 73.70 & - \\ \cline{2-5}
       
       & \multirow{3}{*}{MB-$\mathcal{R}_0$-300} & -   & 77.11 & 2.51 \\
       & & CE  &  77.37 & 2.53 \\
       & & FD~\cite{Yan_2021_CVPR}  &  76.94 & 2.20 \\ \cdashline{2-5}[.7pt/1.5pt]\noalign{\vskip 0.15em}
        
       & \multirow{3}{*}{MB-($\mathcal{R}_2$+CAS)-300} & -  & 76.83 & 2.16 \\
       & & CE & 77.05  & 2.37 \\
       &  & FD~\cite{Yan_2021_CVPR}  & 76.70 & 2.12\\ \cline{1-5}
       \multirow{7}{*}{ResNet} & RN-$\mathcal{R}_0$-2000 & - & 78.25 & - \\ \cline{2-5}
       
        & \multirow{3}{*}{RN-$\mathcal{R}_0$-3000} & - &  78.78 & 2.04 \\
        &  & CE  & 78.68 & 2.47 \\
        & & FD~\cite{Yan_2021_CVPR}  & 78.37 & 1.33 \\ \cdashline{2-5}[.7pt/1.5pt]\noalign{\vskip 0.15em}
        & \multirow{3}{*}{RN-($\mathcal{R}_2$+CAS)-3000} & -  & 78.79 & 1.57 \\
        & & CE & 78.63  & 2.16\\
        &   & FD~\cite{Yan_2021_CVPR}  & 78.47 & 0.96 \\
        \bottomrule
        \end{tabular}
       \end{adjustbox}
      \caption{
      Comparing fine-tuning strategies post inheriting the weights from a super-network. ``-'' denotes no fine tuning, CE denotes cross entropy, and FD denotes focal distillation\cite{Yan_2021_CVPR}. Results averaged across three random seeds.}
    \vspace{-0.5cm}
       \label{tab:finetune}
\end{table}

\subsection{Additional properties of \regnas}
In this section, we study two key properties of \regnas: \textit{transitivity}, as it enables the search for models at diverse flop budgets and the \textit{search direction}.

\begin{figure}[h!]
\centering
    \includegraphics[width=\linewidth]{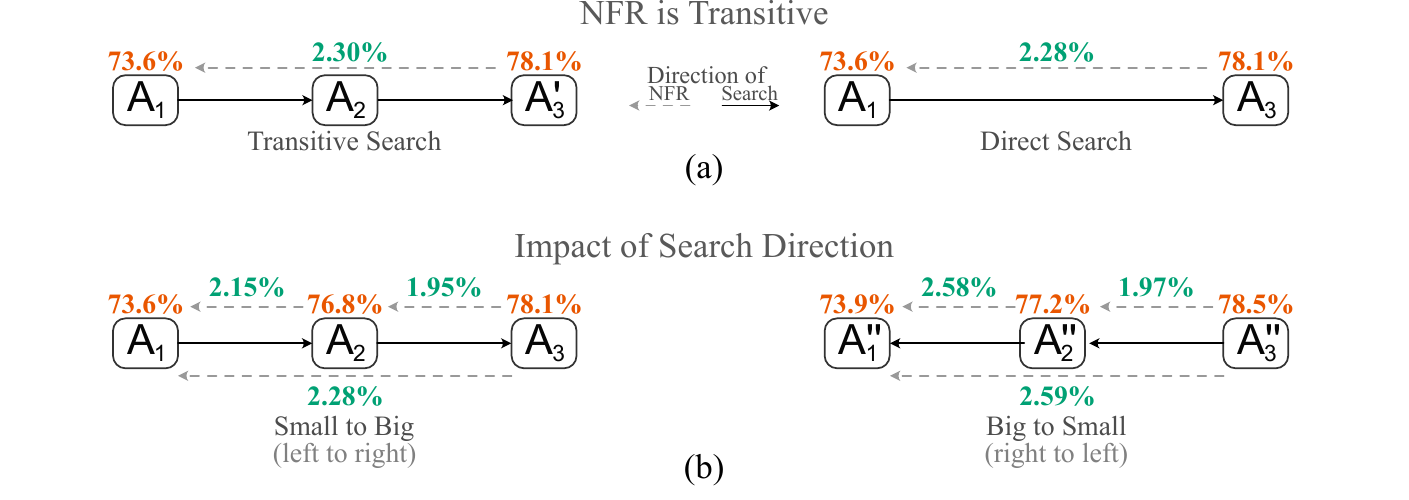}
    \caption{Using the MobileNet-V3 search space of OFA~\cite{OFA} we illustrate two key properties of NFR (a) \textbf{transitivity}: (left) A 600 Mflops model $A_3'$ searched against a 300 Mflops model $A_2$ that was previously searched against a 150 Mflops one $A_1$, automatically achieves a similar NFR as directly searching $A_3$ against $A_1$ (right). (b) \textbf{Effect of search direction}: small to big search (left) leads to lower NFR compared to the opposite direction (right).}
    \label{fig:nfr_properties}
\end{figure}

\medskip
\noindent \textbf{Transitivity of NFR.} 
\label{subsec:transitivity}
Transitivity is measured through the following experiment.
Given a reference model $A_1$, we first search for a model $A_2$ having low NFR using \regnas.
Then we study whether the model $A_3$ searched using $A_2$ as reference model can lead to a low NFR between $A_3$ and $A_1$.
Fig.~\ref{fig:nfr_properties}a illustrates this property
using the MobileNet-V3 search space of OFA~\cite{OFA}.
We can search for a 600 Mflops model in two ways: (1) Transitively search a 600 Mflops model $A'_3$ against a 300 Mflops $A_2$ that was previously searched the 150 Mflops reference $A_1$.  (2) Directly search a 600 Mflops model $A_3$ against the 150 Mflops reference $A_1$; 
Observe that the two models $A_3'$, $A_3$ so obtained achieve a similar NFR with respect to $A_1$, demonstrating that NFR among architectures searched via \regnas is transitive.
Based on transitivity property, when searching for a series of architectures, the $i^{th}$ one can be searched against the $i-1^{th}$ one thereby ensuring a low NFR between the $i^{th}$ and $i-2^{th}$ ones.

\smallskip
\noindent \textbf{Search direction.}
Until now we have focused on a setting in which given a \textit{small} reference model, we search for a \textit{larger} target model. 
We call this the small $\rightarrow$ large setting. Fig.~\ref{fig:nfr_properties}b compares the opposite setting \ie large $\rightarrow$ small. 
We observe that the small $\rightarrow$ large setting benefits in reducing NFR for all sub-nets whereas the opposite direction leads to higher Top-1 accuracy at the cost of higher NFR. 
Since we aim to minimize NFR, we follow the small $\rightarrow$ large setting for all experiments.

\subsection{Generalization Performance of \regnas}
In this subsection, we investigate the performance of \regnas for searching a family of models; from different super-networks and search spaces; with latency constraints on different platforms.

\medskip
\noindent \textbf{\regnas for searching a family of models} 
By leveraging transitive search as discussed in Sec.~\ref{subsec:transitivity}, we can use \regnas to search for a family of models suited for different devices (\eg phone, laptop, cloud).
We demonstrate this by searching for three different models A1, A2, A3 with different flop budgets in Table~\ref{tab:multi_architectures} where we present the NFR between each pair of models. 
Compared to vanilla NAS that optimizes for Top-1 accuracy using $\mathcal{R}_0$ reward, \regnas can consistently reduce the NFR metric.
In some scenarios, the benefit is quite large \eg between models A2, A3 searched in the MobileNet-V3 and ResNet-50 based super-networks of OFA~\cite{OFA}. Please see appendix D for results on four models.

\begin{table}[h!]
    \begin{adjustbox}{width=0.7\columnwidth,center}
    \centering
    \begin{tabular}{ >{\centering\arraybackslash} m{1.5cm} >{\centering\arraybackslash} m{2.0cm} >{\centering\arraybackslash} m{5.8cm}}
    \toprule
    Supernet & Method  & Results \\
    \midrule
    \multirow{2}{*}{MB-V3} & OFA~\cite{OFA} & 
    \includegraphics[valign=b,width=6cm]{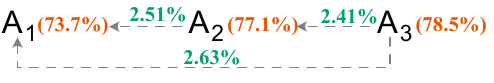} \\
     & \regnas &  
    \includegraphics[valign=b,width=6cm]{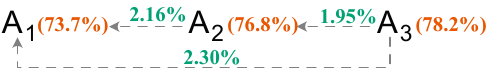} \\
    \cdashline{1-3}[.7pt/1.5pt]\noalign{\vskip 0.15em}
    \multirow{2}{*}{RN-50} & OFA~\cite{OFA} & 
    \includegraphics[valign=b,width=6cm]{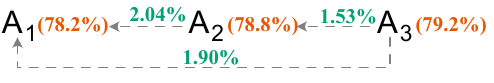} \\
     & \regnas &   
     \includegraphics[valign=b,width=6cm]{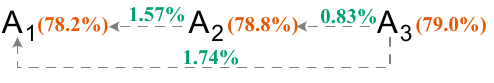} \\
    \cdashline{1-3}[.7pt/1.5pt]\noalign{\vskip 0.15em}
    \multirow{2}{*}{FBNet-V3} & Att-NAS~\cite{AttentiveNAS} & 
    \includegraphics[valign=b,width=6cm]{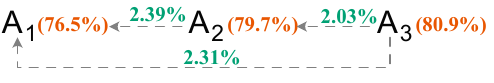} \\
     & \regnas &   
    \includegraphics[valign=b,width=6cm]{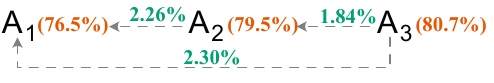} \\
    \cdashline{1-3}[.7pt/1.5pt]\noalign{\vskip 0.15em}
    \multirow{2}{*}{FBNet-V3} &$\alpha$-Net~\cite{alphanet} & 
    \includegraphics[valign=b,width=6cm]{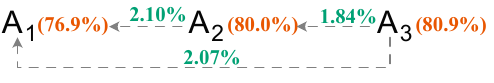} \\
     & \regnas &   
     \includegraphics[valign=b,width=6cm]{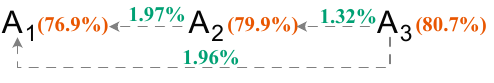}\\
     \cline{1-3}
    \end{tabular}
    \end{adjustbox}
    \caption{
    Testing generalization of \regnas for searching multiple models with diverse super-networks. Model size increases from $A_1$ to $A_3$. NFR is indicated in green and Top-1 accuracy in orange. \regnas successfully reduces NFR in all scenarios.}
    \label{tab:multi_architectures}
\end{table}

\noindent \textbf{\regnas with different supernetworks}
One-shot NAS is an actively evolving research area, with many recent innovations in super-network pre-training.
To demonstrate that \regnas can generalize to different super-networks, we apply \regnas on two state-of-the-art FBNet-V3 based super-networks: AttentiveNAS~\cite{AttentiveNAS} that uses attentive sampling to improve sub-network accuracy and AlphaNet~\cite{alphanet} that uses alpha divergence to improve knowledge transfer from larger to small sub-networks.
Table~\ref{tab:multi_architectures} shows that \regnas consistently reduces the NFR for all pairs of models, with particularly large reduction between models $A_2$ and $A_3$ in the case of AlphaNet.

\begin{table}[h!]
    \centering
    \begin{adjustbox}{width=0.5\columnwidth,center}
    \begin{tabular}{llccc}
    \toprule
    Target  & Model & Latency & Top-1  & NFR (\%)\\
    Device &  & (ms) & $\uparrow$(\%) & $\downarrow$(\%) \\ \midrule
    Note10  &  MB-$\mathcal{R}_0$-15 & 14.83 & 73.34 & - \\ \cdashline{1-5}[.7pt/1.5pt]\noalign{\vskip 0.15em}
    \multirow{3}{*}{1080ti}   & MB-$\mathcal{R}_0$-30 & 29.40 & 77.30 & 2.72\\ \cdashline{2-5 }[.7pt/1.5pt]\noalign{\vskip 0.15em}
    & MB-($\mathcal{R}_2$+CAS)-30 & 29.88 & 77.07 & 2.22 \\\bottomrule
    \end{tabular}
    \end{adjustbox}
    \caption{
    Testing generalization of \regnas under latency constraints. We use a model with 15ms latency on a Samsung Note 10 as the reference to search a target model with 30ms latency on an Nvidia 1080ti GPU. \regnas successfully reduces NFR.}
    \label{tab:Results_edge_to_cloud} 
    \vspace{-0.5cm}
\end{table}

\noindent \textbf{\regnas with latency constraints}
We test whether \regnas can find low NFR sub-networks using latency (instead of flops) as the compute metric.
For this experiment, we search for two models: a reference model with 15ms latency deployed on a Samsung Note 10 phone and a target model with 30 ms latency running on an Nvidia 1080ti GPU.
We measure latency via latency lookup tables built on each device following OFA~\cite{OFA}.
The results in Table~\ref{tab:Results_edge_to_cloud} show that \regnas can indeed find a target model with considerably low NFR without a large drop in Top-1 accuracy while using latency as the compute metric.

\section{Discussion}
In this paper, we tackle the problem of regression (measure via NFR) between models deployed on diverse compute platforms.
We observe that architectures sampled from the same One-Shot NAS super-network naturally share a lot of weights and lead to a low NFR.
Inspired by this finding, we propose a novel architecture constraint that maximizes weight sharing and a novel search reward function which further reduces the NFR.
There are two limitations of our work:
First, the impact of different supernet pre-training strategies on the NFR needs to be investigated.
Our preliminary experiments show that the same super-network (FBNet-V3) obtained via different pre-training strategies (AlphaNet~\cite{alphanet} and AttentiveNAS~\cite{AttentiveNAS}) lead to different NFRs, indicating a measurable impact of super-network pre-training on NFR.
Second, \regnas samples sub-networks from the same super-network which itself may not span the entire range of compute budgets of different platforms \eg tiny embedded platforms to large data-centers.
A future direction may investigate sampling from many different super-networks.
Additionally, designing regression-free models via filter pruning~\cite{duggal2020rest,duggal2021cup} and early-exiting~\cite{duggal2020elf,duggal2021har} are interesting future directions.
Overall, we believe this paper only scratches the surface for an emerging problem---designing regression-free neural networks for diverse compute platforms---that will become more prominent in the era of on-device deep learning.

\clearpage
%
%
\bibliographystyle{splncs04}
\bibliography{main}

\clearpage
\section*{Appendices}

\appendix
\label{sec:appendix}


\section{Implicit reduction of negative flips through weight sharing}
\label{appendix:implicit_nfr_reduction}

\begin{figure*}[h!]
    \centering
    \includegraphics[width=\linewidth]{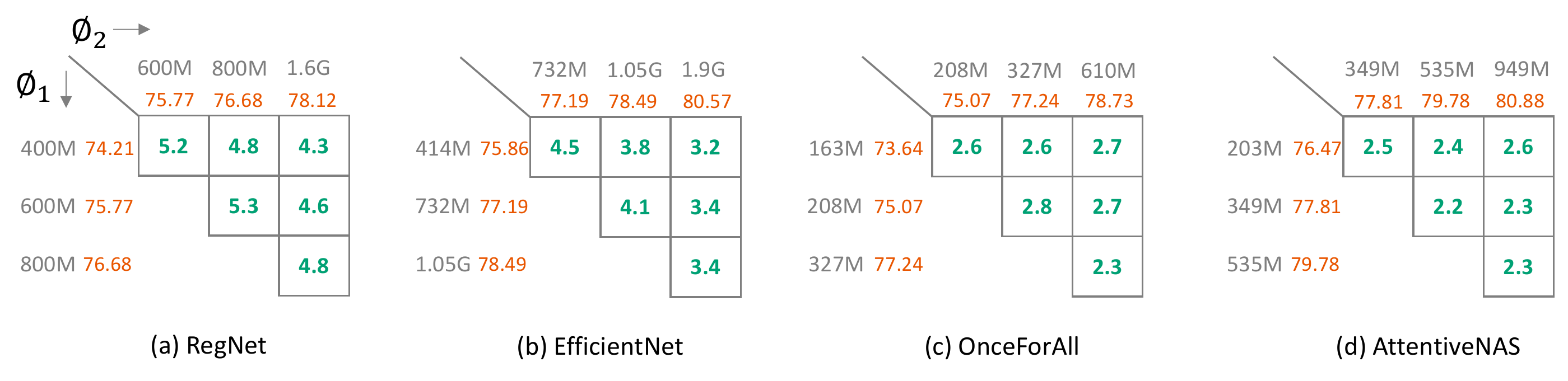}
    \caption{
    We present the Top-1 accuracy (in orange),  pair-wise negative flip rate (in green) and flops (in black) for a family of neural networks obtained via four model design algorithms. The algorithms in (a) and (b) re-train each model independently and lead to much higher NFR compared to One-Shot NAS algorithms in (c) and (d) that jointly train all sub-networks in a super-network. 
    }
    \label{fig:weight_sharing_motivation}
\end{figure*}

We hypothesize that weight sharing leads to lower negative flips. Fig.~\ref{fig:weight_sharing_motivation} presents empirical results to support this hypothesis. 

 Fig.~\ref{fig:weight_sharing_motivation} plots the negative flip rate and Top-1 accuracies from four popular model families---RegNet~\cite{RegNet}, EfficientNet~\cite{EfficientNet}, Onceforall~\cite{OFA} and AttentiveNAS~\cite{AttentiveNAS}. Among these the first two ie. RegNet and EfficientNet independently train the models of different sizes while the latter two i.e. Onceforall and AttentiveNAS sample networks from a common super-network and benefit from weight sharing. From the figure, we can clearly see that the for similar gaps in accuracies, the latter two methods  which benefit from weight-sharing (i.e. Fig.~\ref{fig:weight_sharing_motivation}c,d) lead to much lower negative flip rate than the former two (i.e. Fig.~\ref{fig:weight_sharing_motivation}a,b).

\section{Exploring the Top-1 accuracy \vs NFR tradeoff}
\label{appendix:lambda_ablation}
\begin{table}[h!]
    \centering
    \begin{tabular}{cccc}
        \toprule
        \multirow{2}{*}{$\frac{\lambda_2}{\lambda_1}$} & Mflops & Top-1 Accuracy & NFR   \\ 
        & & $\uparrow$(\%) & $\downarrow$(\%)\\
        \midrule
        0.05 & 297 & 76.91 & 2.67 \\
        0.1 & 295 & 77.09 & 2.70 \\
        0.2 & 298	& 77.03	& 2.63 \\
        0.5 & 299	& 76.94	& 2.27 \\
        1 &	295 & 76.86 & 2.27 \\
        2 & 294 & 76.94 & 2.24 \\
        5 & 290 & 76.74	& 2.15 \\
        10 & 295 & 76.78 & 2.19 \\
        20 & 270 & 76.15 & 2.08 \\
        \bottomrule
    \end{tabular}
    \caption{We measure the impact of varying the weighting factors $\lambda_1$ and $\lambda_2$ on the Top-1 accuracy and NFR of the searched models. Given a fixed 150 Mflop reference model from the MobileNet-V3 search space of OFA, we search a 300 Mflops target model while varying $\lambda_1$ and $\lambda_2$. Observe that large values of $\lambda_1$ prioritize Top-1 accuracy leading to high Top-1, at the cost of high NFR. Large values of $\lambda_2$ prioritizes NFR leading to low NFRs at the loss of low accuracy. Overall, setting $\lambda_1=\lambda_2=1$ leads to reasonable trade-off for both the Top-1 accuracy and NFR.}
    \label{tab:HyperParam}
\end{table}
The search reward of Eq.~4 equally weighs the NFR and Top-1 accuracy by setting $\lambda_1=\lambda_2=1$.
In Table~\ref{tab:HyperParam}, we explore the NFR \vs Top-1 trade-off by setting different values of the $\lambda$ multipliers \ie $\frac{\lambda_2}{\lambda_1} \in [0.05, 20]$.
The results follow the expected trend wherein for $\frac{\lambda_2}{\lambda_1} \in [0.05, 1)$ the Top-1 accuracy is prioritized over NFR which becomes as high as $2.70$. 
At the other end, for $\frac{\lambda_2}{\lambda_1} \in (1, 20]$ the NFR is prioritized over Top-1 accuracy so that NFR becomes as low as $2.08$.
Overall, equally weighing the two metrics, \ie $\frac{\lambda_2}{\lambda_1}=1$ leads to a resonably high Top-1 accuracy with a reasonably low $NFR$.
Thus 
we use $\lambda_1 = \lambda_2=1$ for \regnas.

\section{How significant are the results of \regnas?}
\label{appendix:relative_change}
We strive to find models that achieve a high Top-1 accuracy and low negative flip rate (NFR). 
To demonstrate the significance of our results, we extend Table~1 of the main paper by calculating the
\textit{relative change} of Top-1 accuracy and NFR which is defined as below:
\begin{eqnarray}
 \mathcal{C}(\phi_1, \phi_2) = \frac{\mathcal{M}(\phi_1) - \mathcal{M}(\phi_2)}{\mathcal{M}(\phi_2)},
\end{eqnarray}
where $\mathcal{M} \in \{\text{Top-1},\text{NFR}\}$  and $\phi_1, \phi_2$ are two models.

Table~\ref{tab:main_results} presents the relative change of Top-1 accuracy and NFR between the model searched via \regnas (considered as $\phi_1$) and other models (considered as $\phi_2$).  
The results show that \regnas leads to large relative reduction of NFR \eg upto $50\%$ and $33.5\%$ w.r.t. baseline and state-of-the-art on MobileNet-V3 super-network and upto $67.4\%$ and $48.6\%$ w.r.t. baseline and state-of-the-art on the ResNet-50 super-network of OFA\cite{OFA}.
This reduction in NFR is despite very little (\eg less than $1\%$) relative change of Top-1 accuracy.
These results clearly demonstrate that \regnas can significantly reduce NFR at minimal expense of Top-1 accuracy.

\begin{table}[t]
    \begin{subtable}[h]{0.49\textwidth}
        \begin{adjustbox}{width=0.95\columnwidth}
        \huge
        \begin{tabular}{llcccc}
            \toprule
            & Model  & \multicolumn{2}{c}{Top-1$\uparrow$(\%)}  & \multicolumn{2}{c}{NFR$\downarrow$ (\%)}  \\
            &  &   Abs. & $\mathcal{C}$&  Abs. & $\mathcal{C}$ \\
            \midrule
            ref & MB-$\mathcal{R}_0$-150~\cite{OFA}   & 73.70& & - &\\ \cdashline{1-6}[.7pt/1.5pt]\noalign{\vskip 0.15em}
            baseline  & EffNet      & 77.08 & -0.30 & 4.25 & -49.1\\
            PCT  & EffNet+FD~\cite{Yan_2021_CVPR}  &  76.25 & +0.76 & 3.25 & -33.5\\
            wt. share & MB-$\mathcal{R}_0$-300~\cite{OFA}  & 77.11 & -0.37 & 2.51 & -13.9\\
            proposed &  MB-($\mathcal{R}_2$+CAS)-300  & 76.83 & 0 & 2.16 & 0\\
            \bottomrule
        \end{tabular}
      \end{adjustbox}
      \caption{MobileNet-V3 search space of OFA}
      \label{subtab:main_results_mbv3}
    \end{subtable}
    \begin{subtable}[h]{0.49\textwidth}
    \begin{adjustbox}{width=0.95\columnwidth}
        \huge
        \begin{tabular}{llccccc}
            \toprule
           & Model  & \multicolumn{2}{c}{Top-1$\uparrow$(\%)}  & \multicolumn{2}{c}{NFR$\downarrow$ (\%)}  \\
            &  &   Abs. & $\mathcal{C}$ &  Abs. & $\mathcal{C}$\\
            \midrule
            ref & RN-$\mathcal{R}_0$-2000~\cite{OFA}   & 78.25 & & - &\\ \cdashline{1-6}[.7pt/1.5pt]\noalign{\vskip 0.15em}
            baseline & RN101 & 79.21 & -0.51 & 4.83 & -67.4\\
            PCT     & RN101+FD~\cite{Yan_2021_CVPR}   & 79.90 & -1.37 & 3.06 & -48.6\\
            wt. share & RN-$\mathcal{R}_0$-3000~\cite{OFA}   & 78.78 & -0.02 & 2.04 & -23.0\\
            proposed &  RN-($\mathcal{R}_2$+CAS)-3000  & 78.80 & 0 & 1.57 & 0\\
            \bottomrule
        \end{tabular}
      \end{adjustbox}
      \caption{ResNet search space of OFA}
      \label{subtab:main_results_rn50}
    \end{subtable}
    \caption{Extending Table~1 from the main paper. In addition to the absolute values (abs.), we present the relative change ($\mathcal{C}$) of Top-1 accuracy and NFR.
    EffNet and RN101 represents EfficientNet-B0 and ResNet-101 trained with cross entropy loss while
    EffNet+FD and RN101+FD are trained with state-of-the-art focal distillation loss \cite{Yan_2021_CVPR}.
    Results of weight sharing and our proposed method are averaged from three runs with different random seeds.
    Observe that \regnas leads to a large relative change in NFR at a marginal expense of the Top-1 accuracy. }
     \vspace{-0.5cm}
    \label{tab:main_results}
\end{table}

\section{Extending the search to 4 architectures}
\label{appendix:search_four_archs}
\begin{table}[h!]
    \begin{adjustbox}{width=0.7\columnwidth,center}
    \centering
    \begin{tabular}{  >{\centering\arraybackslash} m{1.0cm} >{\centering\arraybackslash} m{6cm}}
    \toprule
     Method  & Results \\
    \midrule
     OFA~\cite{OFA} & 
    \includegraphics[valign=b,width=6cm]{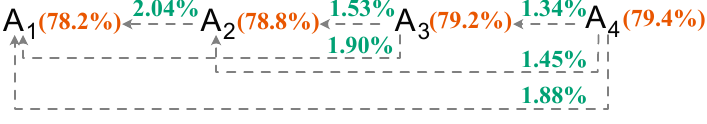} \\
    & \\
     \regnas &   
     \includegraphics[valign=b,width=6cm]{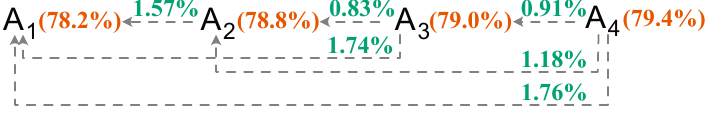} \\
     \bottomrule
    \end{tabular}
    \end{adjustbox}
    \caption{ Testing generalization of \regnas for searching 4 models with diverse compute budgets from the ResNet search space of OFA~\cite{OFA}. Model size increases from $A_1$ to $A_4$. NFR is indicated in green and Top-1 accuracy in orange. \regnas successfully reduces NFR in all scenarios.}
    \label{tab:appendix_multi_architectures}
\end{table}

Table. 4 of the main paper presents the NFR and Top-1 accuracy for three models \ie A1 (2000 Mflops), A2 (3000 Mflops), A3 (4000 Mflops) from the ResNet search space of OFA~\cite{OFA}.
However, one could also extend the search to a larger family of models. 
Table~\ref{tab:appendix_multi_architectures} extends the search to a larger model A4 (5000 Mflops) from the ResNet search space of OFA~\cite{OFA}. 
Observe that the four models A1-A4 searched via \regnas achieve significantly smaller pairwise NFR in all cases, with minimal loss of Top-1 accuracy.
This demonstrates that \regnas generalizes to searching for regression-free models across multiple flops budgets.
Note that for this experiment, we only consider the ResNet space of OFA since the MobileNet-V3 space cannot search sub-networks larger than A3 (600 Mflops).

\end{document}